\documentclass[letterpaper, 10pt, conference]{IEEEtran}
\usepackage{copyright}

\IEEEoverridecommandlockouts

\usepackage[english]{babel}
\usepackage{amssymb}
\usepackage{amsmath}
\usepackage[T1]{fontenc}
\usepackage{cleveref}

\usepackage{url}
\usepackage{graphicx}
\usepackage{latexsym}
\usepackage{algorithm}
\usepackage{algpseudocode}
\usepackage{xfrac}
\usepackage[binary-units]{siunitx}
\usepackage{todonotes}
\usepackage{newclude}
\usepackage{multicol}
\usepackage{tabularx}
\usepackage{booktabs}
\usepackage{threeparttable}
\usepackage{enumitem}
\usepackage{subfigure}

\usepackage[acronym]{glossaries}
\newacronym{icr}{ICR}{Instantaneous Centre of Rotation}
\newacronym{fmcw}{FMCW}{Frequency-Modulated Continuous-Wave}
\newacronym{ro}{RO}{Radar Odometry}
\newacronym{vo}{VO}{Visual Odometry}
\newacronym{svd}{SVD}{Singular Value Decomposition}
\newacronym{ml}{ML}{Machine Learning}
\newacronym{dl}{DL}{Deep Learning}

\definecolor{CommentMatt}{rgb}{1,0.2,0}
\definecolor{CommentDani}{rgb}{0,0,1}
\definecolor{CommentRoberto}{rgb}{0.2,0.8,0.2}
\definecolor{CommentPaul}{rgb}{0.9,0,0}
\definecolor{CommentReview}{rgb}{0.9,0.6,0.2}

\newcommand\numberthis{\addtocounter{equation}{1}\tag{\theequation}}

\crefname{table}{Tab.}{Tabs.}
\crefname{figure}{Fig.}{Figs.}
\crefname{section}{Sec.}{Secs.}
\crefname{equation}{Eq.}{Eqs.}

\usepackage{ifthen}
\newboolean{review}
\setboolean{review}{false} 

\ifthenelse{\boolean{review}}{
\usepackage{ulem}
\newcommand{\OB}[1]{\textbf{Comment:#1}}
\newcommand{\REP}[1]{\textbf{Reply:#1}}

\newcommand{\fix}[1]{{\color{blue}#1}}
\newcommand{\fixS}[1]{{\color{blue}\sout{#1}}}
\newcommand{\fixM}[2]{\fix{#1\marginpar{\color{red}#2}}}
}{
\newcommand{\fix}[1]{#1}
\newcommand{\fixS}[1]{}
\newcommand{\fixM}[2]{#1}
}

\begin{document}

\pagenumbering{gobble} 

\bstctlcite{IEEEexample:BSTcontrol}

\title{
    % \vspace{0.5cm}
	\huge \bf
	What Goes Around: Leveraging a Constant-curvature Motion Constraint in Radar Odometry
}
\author{Roberto Aldera, Matthew Gadd$^\dagger$, Daniele De Martini$^\dagger$, Paul Newman\\
% \thanks{$^\dagger$Equal contribution.}
Mobile Robotics Group (MRG), University of Oxford\\\texttt{\{roberto,mattgadd,daniele,pnewman\}@robots.ox.ac.uk}\\
$^\dagger$Equal contribution
}

% \markboth{
% IEEE ROBOTICS AND AUTOMATION LETTERS. PREPRINT VERSION. ACCEPTED JUNE, 2022
% }%
% {What Goes Around: Leveraging a Constant-curvature Motion Constraint in Radar Odometry}

\maketitle

% notice
\copyrightnotice

\vspace{-1cm}

\begin{abstract}
	This paper presents a method that leverages vehicle motion constraints to refine data associations in a point-based radar odometry system.
	By using the strong prior on how a non-holonomic robot is constrained to move smoothly through its environment, we develop the necessary framework to estimate ego-motion from a single landmark association rather than considering all of these correspondences at once.
	This allows for informed outlier detection of poor matches that are a dominant source of pose estimate error.
	By refining the subset of matched landmarks, we see an absolute decrease of \SI{2.15}{\percent} (from \SI{4.68}{\percent} to \SI{2.53}{\percent}) in translational error, approximately halving the error in odometry (reducing by \SI{45.94}{\percent}) than when using the full set of correspondences.
	This contribution is relevant to other point-based odometry implementations that rely on a range sensor and provides a lightweight and interpretable means of incorporating vehicle dynamics for ego-motion estimation. 
	
% 	This paper leverages vehicle motion constraints to reduce error in radar odometry.
% 	By using the strong prior of how a robot is constrained to move smoothly through its environment, we estimate ego-motion from each landmark and its associated match individually rather than considering all matches at once.
% 	This allows for informed outlier detection of poor matches that are a dominant source of pose estimate error.
% 	By refining the subset of matched landmarks, we see an absolute decrease of \SI{2.15}{\percent} (from \SI{4.68}{\percent} to \SI{2.53}{\percent}) in translational error, approximately halving the error in odometry (reducing by \SI{45.94}{\percent}) than when using the full set of correspondences.
\end{abstract}
\begin{IEEEkeywords}
	radar, sensing, radar odometry, ego-motion estimation, motion constraints, field robotics
\end{IEEEkeywords}

\section{Introduction}
\label{sec:introduction}
% What have we done?
% Why does it matter?
% How is it different from what's gone before?

% I'll leave it here
%\fixM{\label{fix:x.x.x}...}{OBx.x}

\IEEEPARstart{E}{go-motion} estimation using radar-only methods has gained much attention recently in the robotics community.
As a sensor that functions consistently in all weather and lighting conditions, radar has become an attractive alternative to traditional vision and laser-based approaches,
which are of limited use in poor conditions like fog, rain, snow, direct sunlight and shadows, dust, or darkness.
% \fixM{
% \label{5.1.3}
% (with experimental evidence comparing radar to vision in e.g.~\cite{cen2018} and~\cite{hong2020radarslam})
% }{R5.1.3}.
In addition to functioning reliably in diverse environmental conditions, radar has a significantly greater sensing range than its traditional counterparts.
This can be of considerable benefit in enabling earlier vehicle and obstacle detection for improved safety or more robust autonomous navigation for mobile robots in environments where stable static features are sparse.
\fixM{\label{5.1.3}For experimental evidence comparing vision and radar in the ego-motion estimation task, see~\cite{cen2018, hong2020radarslam}.}{R5.1.3}

\begin{figure}
\centering % left, bottom, right, top
\includegraphics[width=0.58\textwidth, trim={0cm 0cm 12cm 0cm},clip]{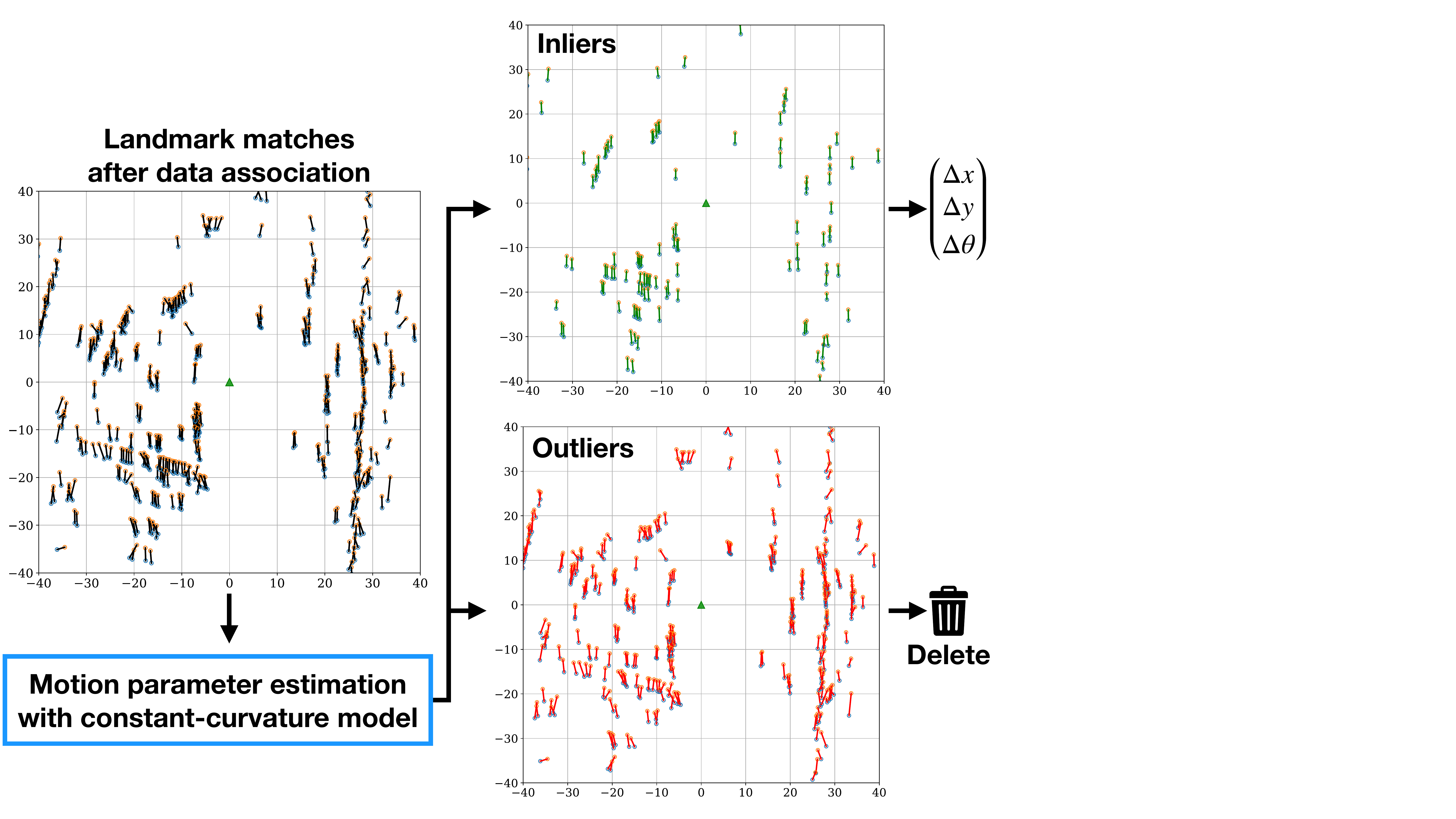}
% \vspace{-.3cm}
\caption{
    The system estimates motion parameters that result from each landmark match.
    These parameters provide an indication of match quality and can be classified as part of the inlier set to keep or the outlier set marked for deletion.
    Odometry accuracy improves when considering only matches corresponding to more plausible ego-motion, as defined by the parameters when under a constant-curvature constraint.
}
\label{fig:sysoverview}
\vspace{-.6cm}
\end{figure}

Building on the foundational \gls{ro} system developed in \cite{cen2018}, this work seeks to leverage a strong prior that previously has not been exploited in this context.
Namely, by acknowledging that the robot's motion can be approximated by an underlying model with certain constraints, we aim to improve the accuracy of ego-motion estimates against some ground truth signal.
%TODO - perhaps explain a tiny bit how we normally use all the matches to do an estimate of overall motion?
This is achieved by developing a model to estimate ego-motion from a single landmark match and expressing this result as a parameter pair.
These parameters fully describe an arc of constant curvature traversed to generate this local observation, allowing each landmark association's odometry estimate to be assessed directly.
Once each match is in this form, it becomes easier to identify associations responsible for introducing significant error into the final estimate, which can naturally be refined based on the best available matches\fixM{
\label{5.2.1}
as shown in~\cref{fig:sysoverview}.
% These are shown as ``deleted'' in~\cref{fig:sysoverview}.
Examples of these poor matches produced by the baseline \gls{ro} system are seen in \cref{fig:matches_problem}, which provides a clearer zoomed-in view.}{R5.2.1}
% \Cref{fig:matches_problem} provides a zoomed-in view with examples of poor matches from our underlying system.

Key contributions in this work are as follows:
\begin{itemize}
	\item The derivation of a model to solve for relative motion in $\mathbb{SE}(2)$ from a single landmark association, as observed with a range-bearing sensor
	\item Comparison of performance between the proposed method, a simpler baseline, and the original system without this additional refinement
	\item Validation of the theoretical contributions on real-world data in complex urban environments~\cite{barnes2020oxford} alongside a comparison with other \gls{ro} systems
\end{itemize}

We proceed by reviewing relevant literature in \Cref{sec:related} and cover background material in \Cref{sec:preliminaries}, before detailing the method in \Cref{sec:method} which illustrates how motion estimates are derived from a single landmark observed twice.
\Cref{sec:experiments} presents the experimental setup used to validate the theoretical contribution of this work, with subsequent results and conclusions covered in \Cref{sec:results} and \Cref{sec:conclusion} respectively.

\section{Related Work}
\label{sec:related}
% Radar work - list a few of the relevant papers, including newer ones, and highlight how they don't leverage a motion model for refinement
%The use of radar as a navigation sensor has recently enjoyed much attention due to its utility in conditions where traditional sensors fail.
The richness of radar data itself has meant a relatively diverse set of approaches to ego-motion estimation and localisation has been developed.
Our system is based on the \acrshort{ro} pipeline of \cite{cen2018} and successive related contributions~\cite{aldera2019fast, aldera2019itsc}. 
Learnt approaches to the odometry task have included an end-to-end method from Barnes et al. \cite{barnes2020masking}, which uses the cross-correlation between scans to predict pose after removing distractor features, as well as a point-based method that provides greater interpretability \cite{barnes2020under}.
Burnett et al. looked at the effects of failing to correct errors in landmark positions resulting from motion distortion at higher speeds \cite{burnett2021we} and later presented a hybridised \acrshort{ro} framework \cite{burnett2021radar} that combines probabilistic trajectory estimation with learnt features to benefit from both classical and data-driven approaches.

Other work that has made use of radar as a navigation sensor includes that by Park et al.~\cite{park2020pharao} which applies the Fourier-Mellin Transform to log-polar images computed from downsampled Cartesian images, and Kung et al.~\cite{kung2021normal} which uses a normal distribution transform typically applied to 2D and 3D LiDAR.
Adolfsson et al.~\cite{adolfsson2021cfear} employ filtering to retain the strongest azimuthal returns and compute a sparse set of oriented surface points, while Hong et al. \cite{hong2020radarslam} use vision-based features and graph-matching in a radar context.

However, previous work on radar-only ego-motion estimation has directly attempted to leverage the dynamic constraints of the system in operation.
This is a natural consideration, as exploiting the underlying smoothness in real trajectories traversed by sensors mounted on non-holonomic robots has proven useful in a wide range of contexts in the past.
Using cubic B-splines, Lovegrove et al.~\cite{lovegrove2013spline} looked at leveraging the torque-minimal motion of their visual-intertial system approximated by a parameterised spline, built using relative poses as control points to describe a continuous pose estimate on an interval between discrete poses.
This effectively constrains the underlying motion but is not directly applicable to online ego-motion estimation when future pose estimates are unavailable.

Tong et al.~\cite{tong2013gaussian} developed Gaussian Process Gauss-Newton as a non-parametric state estimation framework where similarly to \cite{lovegrove2013spline}, the continuous-time reality of robot motion is acknowledged when processing measurements.
Although our work does not directly impose a continuous-time framework, we do leverage the continuous nature of our robot's trajectory when assuming that it moves smoothly between poses.

As a range-based sensor, radar is in some respects similar to 2D LiDAR, making the work in \cite{almeida2013real} of interest where authors used a non-holonomic extended Kalman Filter motion model to refine laser scan match estimates.
Earlier, the authors in~\cite{nourani2009practical} proposed a \gls{vo} system that solved for Ackerman model parameters directly from feature pixel displacement using a ground-facing camera.
However, these methods \textit{enforce} a motion constraint to either directly estimate or smooth the odometry instead of \textit{leveraging} the underlying model to isolate poorer matches as a refinement step.

Scaramuzza et al.~\cite{scaramuzza2009real} present 1-point RANSAC which draws on this idea of exploiting a vehicle's motion constraints in an odometry system and is perhaps the closest work to our contribution, but is only applicable in the vision context.
They show how outliers can be discarded in monocular odometry by using a restrictive motion model.
However, the details between their work and that which is presented here differ significantly due to \fixM{\label{5.1.1}the sensor measurement model -- radar is a range sensor that measures distances directly, unlike the vision context.
We solve this particular problem for range-azimuth sensors, with the full derivation in the appendix.
Furthermore,~\cite{scaramuzza2009real} do not use a baseline odometry system to propose their initial correspondences, as we do.
Lastly, empirically we found the histogram voting used in~\cite{scaramuzza2009real} performs poorly in this domain and so present a new selection approach using the lower and upper quantiles of the estimated motion parameter.
More details of these differences follow in~\cref{sec:method}.}{R5.1.1}

% the fundamental differences between sensor modalities.

\section{Preliminaries}
\label{sec:preliminaries}
We use a Navtech \acrshort{fmcw} scanning radar which rotates about its vertical axis and continuously senses the environment using frequency-modulated radio waves.
% During each rotation, the sensor inspects one \textit{azimuth} ($\alpha$) at a time and receives a power signal that is a function of reflectivity, size, and orientation of objects along the specific azimuth and at a particular distance, $r$.
Measurements are captured along an azimuth ($\alpha$) at one of $M$ discrete angular positions and return a single power reading (a function of reflectivity, size, and orientation of objects) for $N$ range \textit{bins} corresponding to distances, $r$.
We refer to one full rotation through all $M$ azimuths as a \textit{scan} $\mathcal{S}$.
Furthermore, let $\mathbf{s}(k) \in \mathbb{R}^{N{\times}1}$ be the power-range readings at time step $k$, where $t(k) = t_k$ is the time value at $k$ and $\alpha(k) \in A$ is the azimuth associated 
with the measurement.
The element $s_i(k) \in \mathbf{s}(k)$ is the power return at the $i$-th range bin, with $i \in \{1, \ldots, N\}$; its measurement range is given by $r_i(k) = \beta(i - 0.5)$, where $\beta$ is the range resolution.
% RA: don't use \rho for range to a landmark, it's used later as the distance between poses, and r0 and r1 are the ranges to the landmark.

Our contribution builds upon the \gls{ro} pipeline first described in~\cite{cen2018}, which we summarise here.
Firstly, landmark features are extracted from each azimuth's power-range spectrum captured by the radar as it scans the environment.
Although this process is not detailed here, it is well known to be difficult in practice due to the complexity of radar measurements themselves, which may often be obscured by multipath reflections, harmonics, speckle noise, and other hindering effects~\cite{robo_radar}.

To compute odometry from the extracted features in two successive landmark sets $\mathbb{L}_{t_k}$ and $\mathbb{L}_{t_{k-1}}$, landmarks in $\mathbb{L}_{t_k}$ must be associated with a single corresponding landmark in $\mathbb{L}_{t_{k-1}}$.
We pose this data association as a pairwise problem as in \cite{leordeanu2005spectral}, postulating that the distance from each point to its neighbours as observed in one instance uniquely describes that point in any proximal observation, regardless of the relative translation and rotation between observations.
This assumption allows landmarks to be matched to counterparts in any other landmark set containing sufficient overlap.
Once unary candidates have created preliminary associations and final matches selected using the largest elements from the principal eigenvector of the pairwise compatibility matrix (see \cite{cen2018,aldera2019itsc} for more detail), a final pose estimate $G_{t_{k-1}, t_k}$ is computed by a \gls{svd} algorithm \cite{challis1995procedure}.
\fixM{\label{5.4}For a graphical representation of this \gls{ro} pipeline, please refer to Fig. 5 on page 5 of Cen and Newman~\cite{cen2018}.}{R5.4}

Our contribution interfaces with the system components described above, taking these associated landmarks as inputs.
We seek to refine the association and pose estimation process by considering the physical constraints of the robot platform and leveraging this additional information to improve odometry accuracy.

\begin{figure}
\centering
\includegraphics[height=9cm,keepaspectratio]{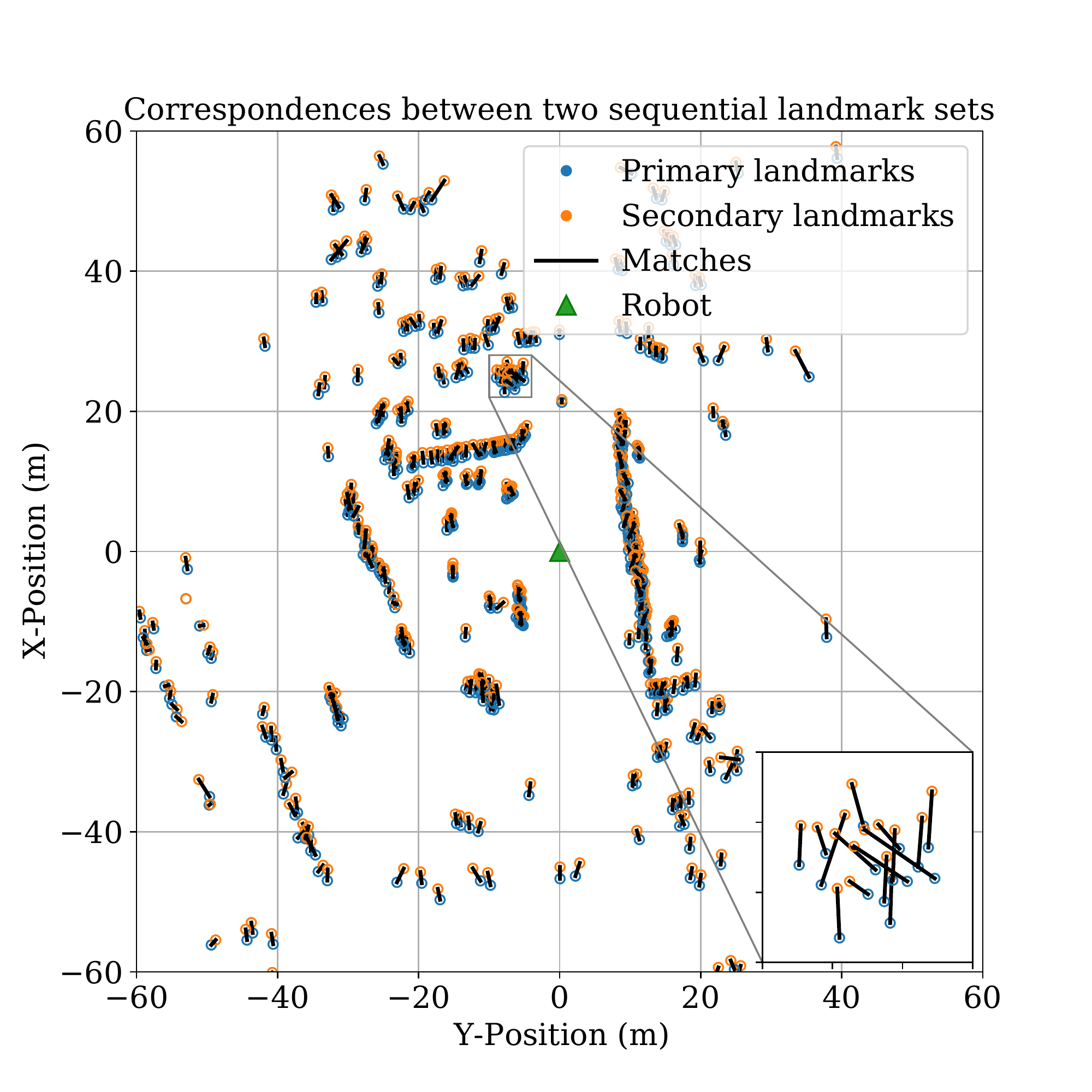}
% \vspace{-5mm}
\caption{
Two successive sets of associated radar landmarks are shown from a scan match, cropped from the full radar range of \SI{165}{\meter} to \SI{60}{\meter} for clearer visualisation.
After observing the orange landmarks, the robot shown at the origin as a green triangle moved straight ahead by roughly \SI{1.4}{\metre} and then observed the blue landmarks.
Although most matches indicate this motion relatively clearly, there are few stark examples where associated points imply contradictory relative motions, which contribute disproportionately to ego-motion estimate error.
}
\label{fig:matches_problem}
\vspace{-5mm}
\end{figure}

\begin{figure*}%[!h]
\centering
\includegraphics[width=16cm,keepaspectratio]{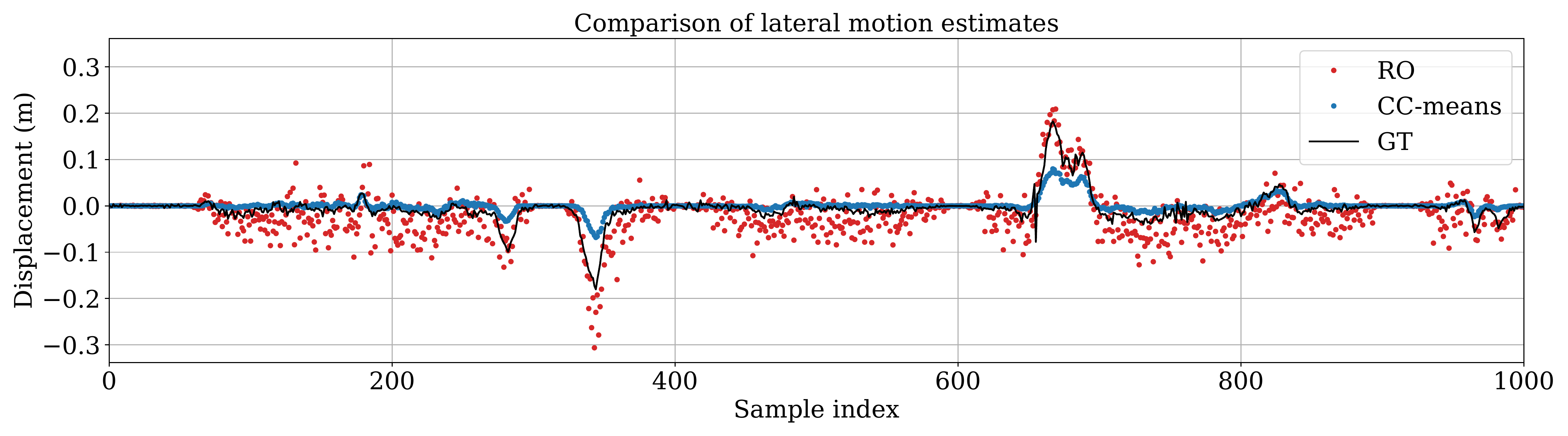}
% \vspace{-5mm}
\caption{
Without leveraging any knowledge of platform constraints, pose estimates from \gls{ro} in red tend to produce relatively poor performance in lateral motion estimation, particularly evident when driving straight, where estimates deviate by up to as much as \SI{10}{\centi\metre}.
Our contribution shown in blue produces smoother estimates in these $\Delta y$-values which are a nearer approximation of the ground truth signal in black.
}
\label{fig:lateral_motion_problems}
\vspace*{-5mm}
\end{figure*}

\Cref{fig:matches_problem} illustrates the problem we have identified where it is clear that despite the best efforts in landmark extraction and data association, poor matches will persist.
\fixM{\label{2.2.3}Those shown here are from the baseline system after the data association step, which is tasked with selecting the best candidates.
In this approach, a point is associated with the most similar available point based on distances and angles to other points (i.e the scene geometry), which does not consider vehicle kinematic constraints.
At first it appears as if an outlier rejection scheme like RANSAC would explicitly remove these outliers, however we show in~\cref{sec:results} that a standard application of RANSAC only marginally improves ego-motion estimation accuracy, as it also does not exploit domain knowledge of how the vehicle moves.}{R2.2.3}
The effect of failing to remove these poor matches is shown in \cref{fig:lateral_motion_problems}, where implausible lateral motion is estimated due to poor correspondences.

\section{Method}
\label{sec:method}
This section details approaches to mitigating problematic matches in \cref{fig:matches_problem}, outlines the proposed model for leveraging a constant-curvature constraint, and describes pose estimation from a refined subset of candidates.
\fixM{\label{2.1.1}We assume the motion of our vehicle is described by a non-holonomic kinematic motion model, appropriate to two-wheel drive urban vehicles which traverse local arcs of constant curvature between two successive measurements, detailed in~\Cref{subsec:motion-model}.}{R2.1.1}
% The assumption core to this approach is a non-holonomic kinematic motion model appropriate to two-wheel drive urban vehicles which we use to describe the vehicle's motion as on arcs with constant curvature, detailed in~\Cref{subsec:motion-model}. 

\subsection{Outlier rejection baseline}
\label{subsec:outlier-baseline}
% As a first pass, we try and select the best associations using an outlier rejection strategy - RANSAC in this instance can be used to filter out landmark associations that are beyond a certain threshold (matchstick diagram here may help a lot) - so inliers are considered as matches that are close to the particular model's SVD solve (short error distances, with range compensation)
Before leveraging a constant-curvature constraint, we first consider a sensible starting approach that provides a suitable baseline.
One instinctive way to improve pose estimate accuracy is to identify and discard poor associations.
Matches contributing disproportionately to the final pose error can be flagged as outliers by a simple RANSAC \cite{fischler1981random} implementation before the \acrshort{svd} calculation.
By considering landmark set $\mathbb{L}_{t_{k-1}}$ in the reference frame of $\mathbb{L}_{t_k}$ (which amounts to transforming $\mathbb{L}_{t_{k-1}}$ into the same frame using the computed pose estimate $G_{t_{k-1}, t_k}$), associations can be assessed individually.
Practically, RANSAC is run by randomly selecting two matched landmarks and computing a pose estimate with the \acrshort{svd} from this minimal subset.
All other candidates are then classified as inliers or outliers based on their range-compensated distance away from their associated partner when under the proposed pose estimate.
The sampled model that produces the most inliers after a fixed number of iterations yields the best proposal.
% RA: leaving this RANSAC formula out, to avoid readers getting caught up thinking we use RANSAC in our final method.
% As per \cite{fischler1981random}, we calculate the number of iterations $k$ required to ensure with a desired probability $p$ that one of the trials contains only inliers when the probability of selecting an inlier for each of the $n$ candidates required is $w$ using: $$k = \frac{log(1-p)}{log(1-w^n)}$$
% A new larger subset containing the maximum number of inliers is then passed to the \acrshort{svd} method for a final pose estimate.
These inliers are passed to the \acrshort{svd} method for a final pose estimate.
This serves as a baseline to compare our method against, which takes a more informed approach that does not use RANSAC.
% This is useful because it provides a baseline of expected performance if we reject poor matches, but without explicitly using a motion model (our strong prior)

\subsection{Constant-curvature motion constraints}
\label{subsec:motion-model}
% Inspired by 1-point RANSAC, we develop a model to describe ego motion from a single landmark match along a portion of a circular arc, a special case of clothoid that underlies the smooth motion our robot likely moved along between measurements (describe method here in greater detail including diagram showing how from 2 ranges and bearings a pose can be fixed to a single solution). Also mention corner cases, like straight motion or when the robot is stationary.
Inspired by \cite{scaramuzza2009real}, we seek to describe ego-motion using only a single landmark association.
However, in contrast to~\cite{scaramuzza2009real}, our formulation does not use pixel coordinates and epipolar geometry required in vision, but instead develops the corresponding theory for range sensors that provide range and azimuth readings.
\fixM{\label{5.1.2}To this end, we present a similar yet novel expression in eq.~(A.7) in the appendix for the constraint parameter to be estimated, described below.}{R5.1.2}

\begin{figure}
\centering % left, bottom, right, top
\vspace{.5cm}
\includegraphics[width=6.5cm,height=6.5cm,keepaspectratio,trim={24cm 4.5cm 9cm 3cm},clip]{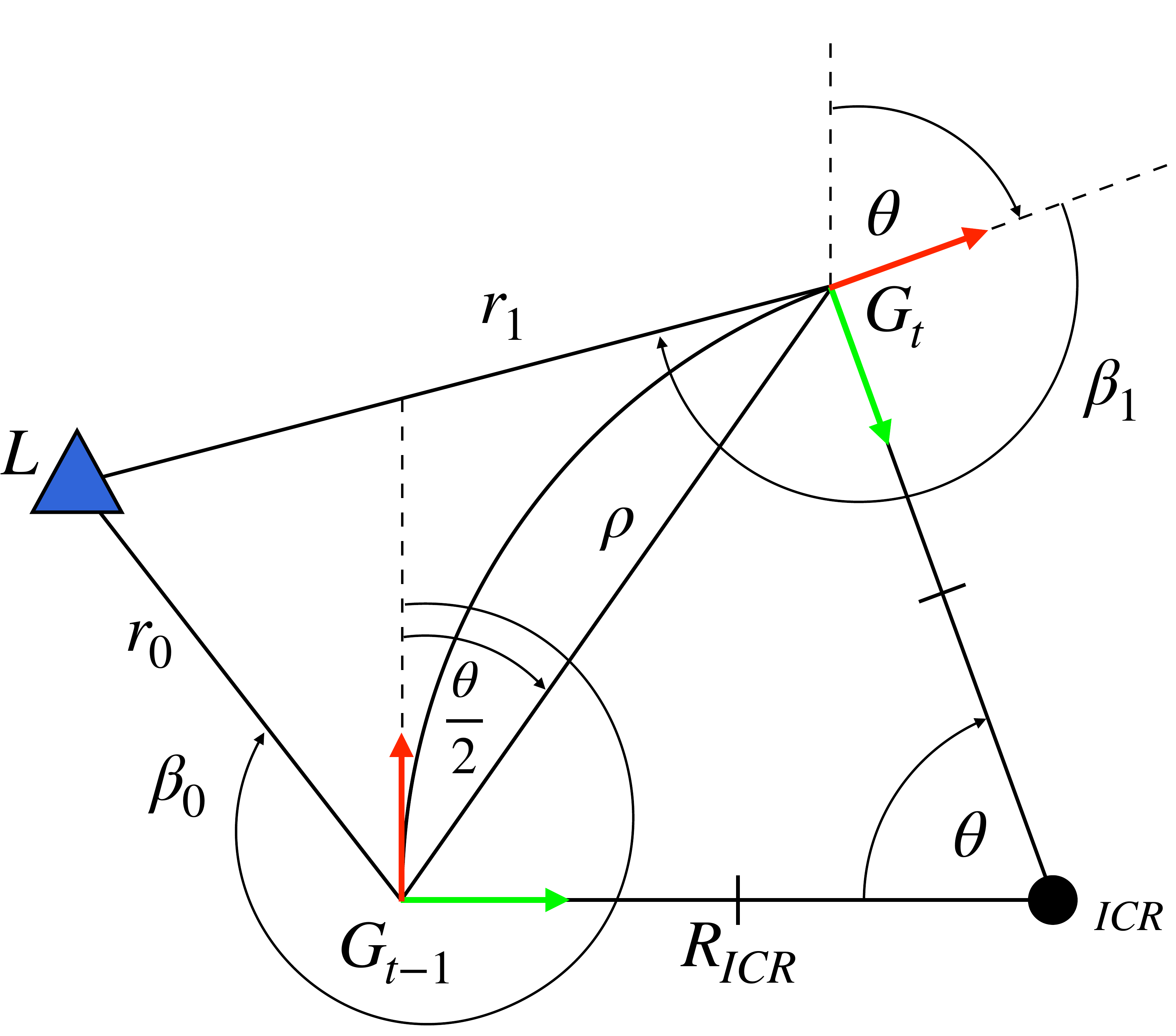}
\caption{
% A landmark at $L$ is observed from the robot at pose $G_0$ and then again at $G_1$ having moved along a circular arc.
% Relative ranges and bearings are used to determine the relative motion between these two observations.
Associated observations of a landmark $L$ from the robot at pose $G_{t-1}$ and then at $G_t$ can be used to describe the local relative motion as a circular arc.
}
\label{fig:cm-diagram}
\vspace{-.5cm}
\end{figure}

When matching two landmark sets to compute a relative pose in $\mathbb{SE}(2)$, at least two pairs of associated landmarks need to be available to constrain the solution.
With only a single pair, the relative motion cannot be fully resolved as there would be infinitely many poses on a circle around that landmark that satisfy the single constraint.
However, by leveraging a constant-curvature constraint to describe the local robot motion between poses, it is possible to resolve relative motion using only one landmark association pair.
% To this end, as in~\cite{scaramuzza2009real}, we employ a circular motion model to describe an arc of constant curvature along which the robot is constrained to move locally.
%, as a local approximation to the underlying clothoid path traversed. 
%\gadd{-- transition curves used in the construction of roads}
%\gadd{citation needed}  %RA - I'm not sure we need a citation here, it's pretty intuitive that car-like robots have to move on a curved path because they have a steering wheel, so any traversal must have continuous curvature (and we just simplify that and say it's constant over a quarter second interval)
%\fixM{\label{fix:x.x.x}...}{OBx.x}
This model serves as a suitable approximation for the motion of \fixM{\label{2.1.3}the vehicle in the experimental dataset (see~\cref{sec:results})
which was captured in an}{R2.1.3} urban environment where wheel slippage is minimal.
\fixM{\label{2.1.2}Additionally, the short \SI{0.25}{\second} duration between successive scan measurements means that the curvature of the traversed path can be assumed to be constant}{R2.1.2}.
\Cref{fig:cm-diagram} shows two successive robot poses where an arc has been traversed along the circumference of a circle centred at the \gls{icr} of radius $R_{ICR}$, through an angle of $\theta$.
\fixM{\label{2.1.4}The constant curvature constraint here is applied only between these two poses at which consecutive scans are collected, and does not consider a window of scans or poses before this current estimate.}{R2.1.4}
The final pose is a distance $\rho$ from the starting position, with a yaw offset of this same $\theta$.
Relative to the starting pose, the final pose is at an angle of $\sfrac{\theta}{2}$ in this case, observing that both poses are at a distance $R_{ICR}$ from the \gls{icr}, forming an isosceles triangle containing $\theta$ and two other equal angles.
For straight trajectories, the robot can be described as moving along a circle of infinite radius.

\subsection{Pose estimation from a single landmark association}
\label{subsec:single-landmark}
\begin{figure}
\centering
\vspace{.5cm}
\includegraphics[height=6.5cm,keepaspectratio,trim={2cm 4cm 10cm 2cm},clip]{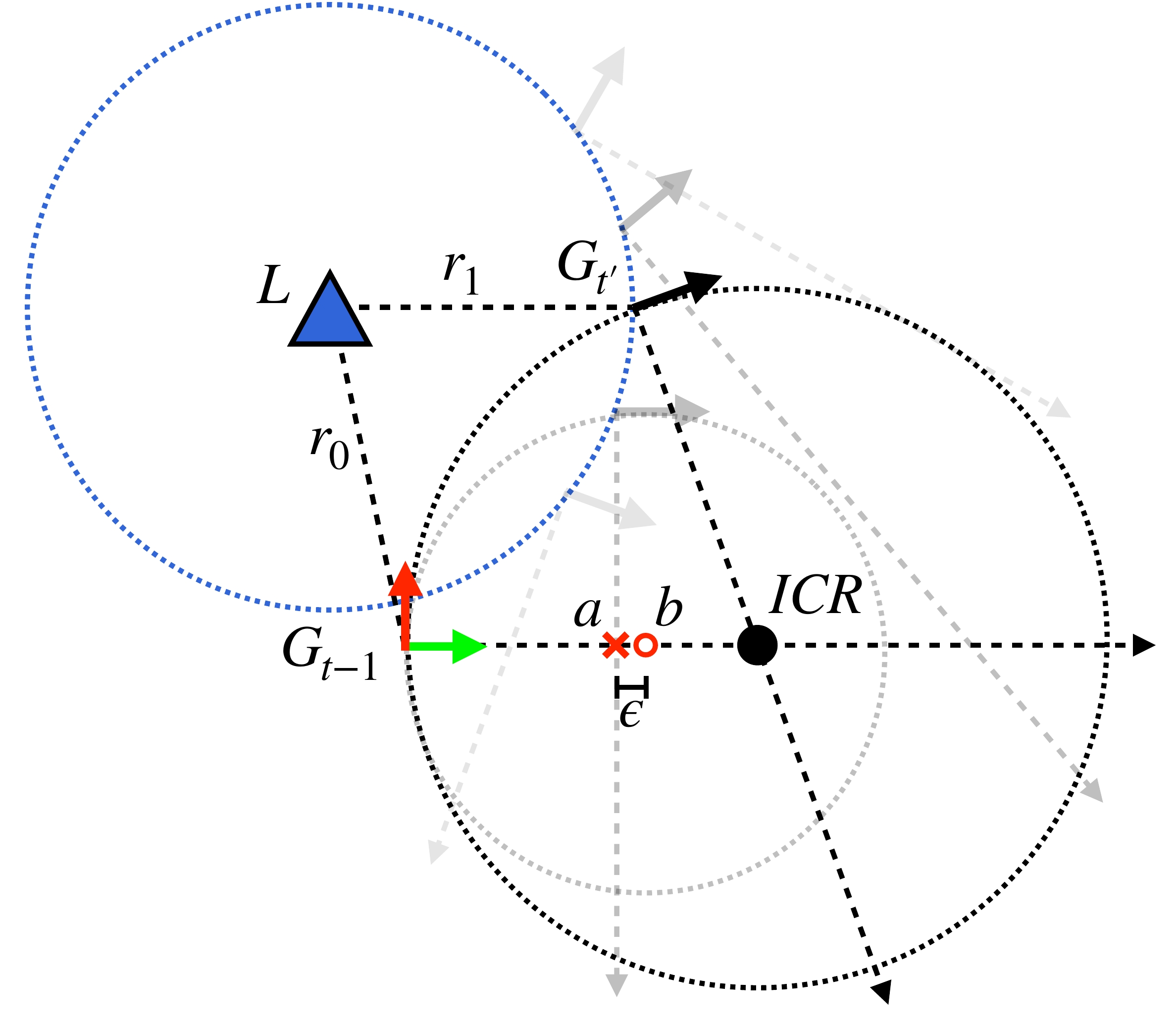}
\caption{
Once a landmark $L$ has been observed initially from pose $G_{t-1}$ to be at a certain range and bearing angle, the task is now to find where $G_{t'}$ is relative to $G_{t-1}$ given a new range and bearing to that same associated landmark from the new pose.
This diagram illustrates that although there are infinitely many possible $G_{t'}$ poses that satisfy the second range and angle measurement around a locus, which is a circle centred on $L$ of radius $r_1$, only one of them will satisfy the constant-curvature motion constraint too.
Notice how for the pose slightly below the correct $G_{t'}$ where a faded proposed circle has been drawn, there is an error $\epsilon$ between the point $a$ at which their extended axes meet and the centre of that circle $b$.
These points must coincide if this $G_{t'}$ guess were to fully satisfy both the landmark observation and the constant-curvature motion constraints.
}
\label{fig:single-landmark}
\vspace{-.5cm}
\end{figure}

A single landmark observed from the starting pose is at a range $r_0$ and bearing $\beta_0$.
At the next instance, once the robot has reached the final pose, that same landmark is observed again, now at a range $r_1$ and bearing $\beta_1$.
These four parameters allow us to express the robot's motion along the arc that subtends $\theta$ from the \gls{icr} where $\theta$ and $R_{ICR}$ are expressed as our constant-curvature motion parameters:
\begin{gather}
	\theta = 2\arctan\left(\frac{\sfrac{r_0}{r_1}\sin \beta_0 - \sin \beta_1}{\sfrac{r_0}{r_1}\cos \beta_0 + \cos\beta_1}\right)\\
	R_{ICR} = \frac{r_1\sin(\beta_0 - \beta_1 - \theta)}{2\sin\left(\sfrac{\theta}{2}\right)\sin\left(\sfrac{\theta}{2} - \beta_0 \right)}
\end{gather}
And from these parameters, $\Delta x$, and $\Delta y$ are simply
\begin{align}
	\Delta x &= \rho \cos \left(\sfrac{\theta}{2}\right) \\
	\Delta y &= \rho \sin \left(\sfrac{\theta}{2}\right) 
\end{align}
where $\rho = 2R_{ICR} \sin \sfrac{\theta}{2}$ and $\Delta \theta$ is the same as its constant-curvature parameter counterpart.
The reader is directed to the appendix for a full derivation.
Note that, for the edge case when driving in a perfectly straight line, $\theta=0$ and $R_{ICR}$ is infinite, as mentioned in \Cref{subsec:motion-model}.
Although theoretically possible, this tends only to happen in practice when $\beta_0 = \beta_1$ and $r_0 = r_1$ which occurs when the robot is stationary, where we set $\Delta x = \Delta y = \theta = 0$.
\Cref{fig:single-landmark} attempts to illustrate intuitively how the combination of a single associated landmark and this constant-curvature motion constraint is all that is required to solve for a relative pose estimate. % that satisfies the range and bearing observation, while moving along a circular arc.

\subsection{Final pose estimation using a conforming subset}
\label{subsec:final-pose-estimation}
Once each associated landmark match is expressed in terms of $\theta$ and $R_{ICR}$, those that yield $\theta$-values near the median generally correspond to matches that have captured the underlying ego-motion.
Associations that produce more extreme $\theta$-values usually correspond to dynamic objects or spurious radar power returns that were similar enough between two observations to pass through the initial data association algorithm.
Unlike in \cite{scaramuzza2009real}, we found in this context that simply using the median $\theta$ association to calculate the final pose estimate led to inconsistent estimation performance.
\fixM{\label{5.1.4}It appears in~\cite{scaramuzza2009real} that $\theta$-values are discretised, with the middle of the most populated bin taken as the solution.
In practice, we found in early experiments that the histogram approach performed poorly in this context.
We instead choose to retain a conforming subset and avoid any discretisation errors which may arise.}{R5.1.4}
We also experimented with considering the $R_{ICR}$ values when evaluating candidates but found they did not give a strong indication of a candidate's credibility.
Based on these observations, we select a subset of matches when calculating the final pose estimate, which can be done by sorting all $\theta$-values in ascending order and taking an inter-quantile set, where specific quantiles to use are left as a hyperparameter.
The final pose estimate is produced by passing every match in the selected subset to the \acrshort{svd} algorithm as before.

\begin{figure}
\centering
\includegraphics[width=\columnwidth,keepaspectratio]{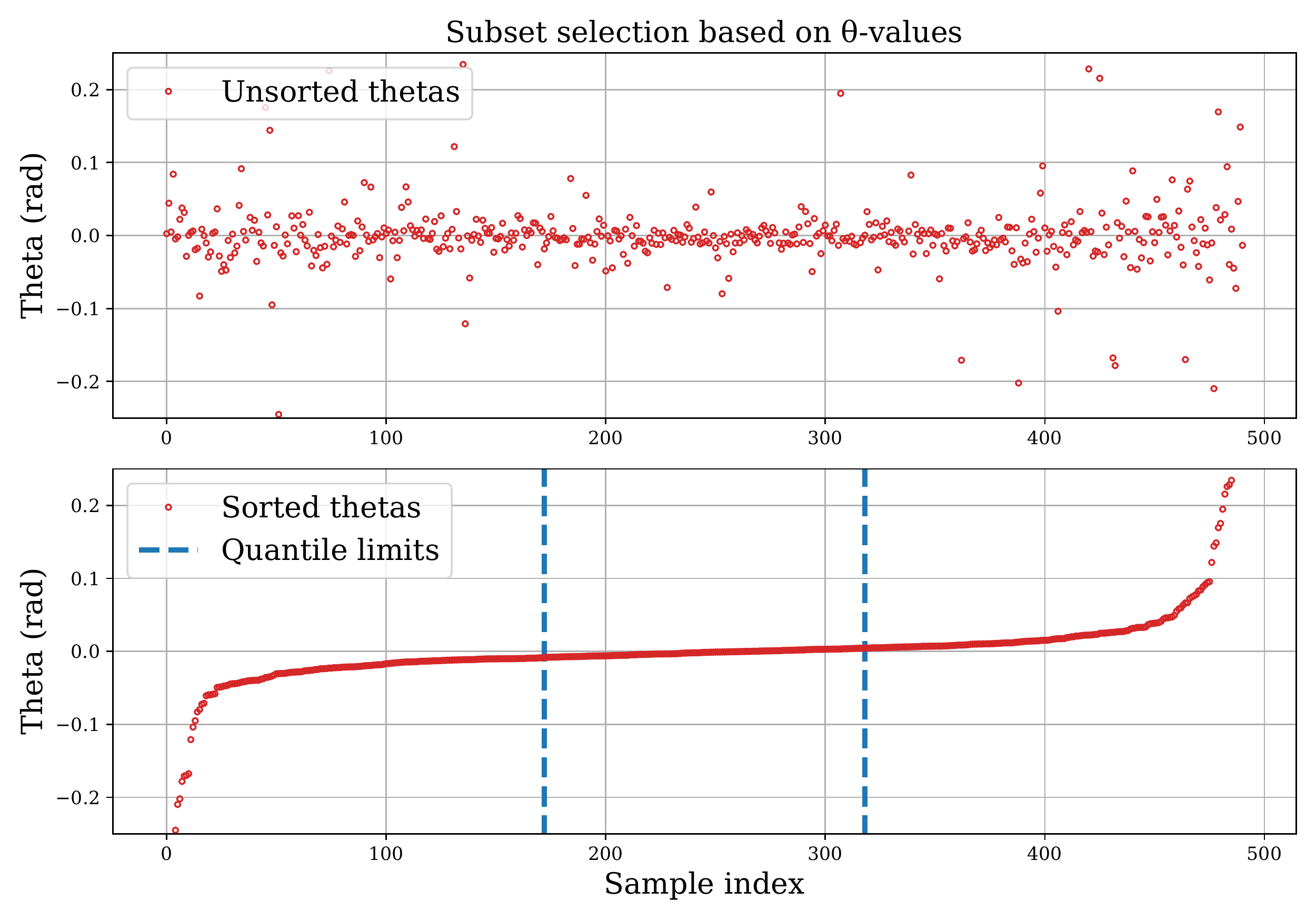}
\caption{
        After calculating a $\theta$-value for each individual match between landmark sets from adjacent scans, we can sort them to get an indication on what the majority of matches are estimating.
        Poor matches tend to produce more extreme $\theta$-values that do not align with the consensus.
        By taking a subset between two quantiles, we are able to exclude these less reliable (and in some instances more obviously erroneous) matches when computing the final pose estimate.
        }
\label{fig:sorted-thetas}
\vspace{-.5cm}
\end{figure}

\Cref{fig:sorted-thetas} shows how when these $\theta$-values are sorted they tend to form a general consensus, with outliers being pushed to the extremities.
However, we found that estimates improve if we instead produced $\Delta x$, $\Delta y$, and $\Delta \theta$ estimates for individual matches in the subset and selected the mean of each of these parameters as our final pose estimate.
Note that although we rely on our motion model to assess the reliability of each landmark association, the final estimate is not forced to conform to these constraints and instead is free to describe more complex and realistic robot motions.
% constrained under this same schema.

%Landmark correction, pointnet, all that jazz... -> separate for now
%\subsection{Correcting landmark positions using 1-match circular motion estimate framework}
%\label{subsec:landmark-correction}
%Give it to a pointnet.

\section{Experimental Setup}
\label{sec:experiments}
To evaluate our proposed system, we utilise the \textit{Oxford Radar RobotCar Dataset} \cite{barnes2020oxford} as its optimised ground truth signal is required to assess if \gls{ro} accuracy has been improved.
Our experimental data consists of the first five sequences of the dataset, totalling approximately \SI{50}{\kilo \metre} of driving.
Landmarks are extracted and associated as previously described in \cite{cen2018} prior to our proposed refinement process.
The use of this \acrshort{ro} framework is convenient as it allows us to build on a familiar system, but we emphasise that our contribution is transferable to other point-based odometry implementations that use measurements from a range sensor.
% it should be noted that other suitable landmark extraction and association methods could be employed.

\subsection{RANSAC baseline}
\label{subsec:ransac-baseline}
As discussed in \Cref{subsec:outlier-baseline}, we implement a rudimentary RANSAC-based outlier rejection system as a baseline.
% to remove matches that are not counted as inliers by the best-performing model.
This provides a reference point in assessing performance when matches have undergone some form of refinement but without leveraging a constant-curvature constraint.

\subsection{Selecting a subset using constant-curvature parameters}
\label{subsec:subset-selection}
When determining which matches to consider in the final pose estimation step described in \Cref{subsec:final-pose-estimation}, we found using $\{0.35, 0.65\}$ as the quantile limits yields satisfactory performance for this dataset, slightly more so than using the first and third quartiles.
These may need to be adjusted when operating in settings with a different distribution of misleading associations, like in open rural environments where there may be fewer reflections or scenarios with an increased number of dynamic objects in the scene.

\subsection{Processing the selected subset to produce pose estimates}
\label{subsec:producing-pose-estimates}
After landmark associations have been reduced to a conforming subset, \Cref{subsec:final-pose-estimation} described two methods to compute a final pose estimate.
Final poses are estimated using both the \acrshort{svd} method as well as the simpler approach of taking the means of $\Delta x$, $\Delta y$, and $\Delta \theta$ to illustrate relative performance.
% , alongside the result of using all of the original the proposed matches.
% TODO?: \roberto{Not sure about this, it might be easier to just show SVD isn't great once and leave it at that.}

\subsection{Performance metrics}

For evaluation we follow the KITTI odometry benchmark~\cite{geiger2012we} which evaluates the distance-normalised translational and rotational error over all possible fixed segment lengths ranging from \{\SI{100}{\meter}, \SI{200}{\meter}, ..., \SI{800}{\meter}\}.

This approach is commonly used when assessing odometry performance, as it provides a balanced metric that unlike end-point errors is not overly sensitive to \textit{when} in the sequence an error occurs, i.e. an error in a rotational estimate near the start would tend to produce higher end-point errors than if occurs near the end.
Our evaluation is based on the implementation provided by the STARS laboratory\footnote{Evaluation metrics implementation available here: \url{https://github.com/utiasSTARS/pyslam}}.

\begin{figure}
\centering
\includegraphics[width=\columnwidth,keepaspectratio]{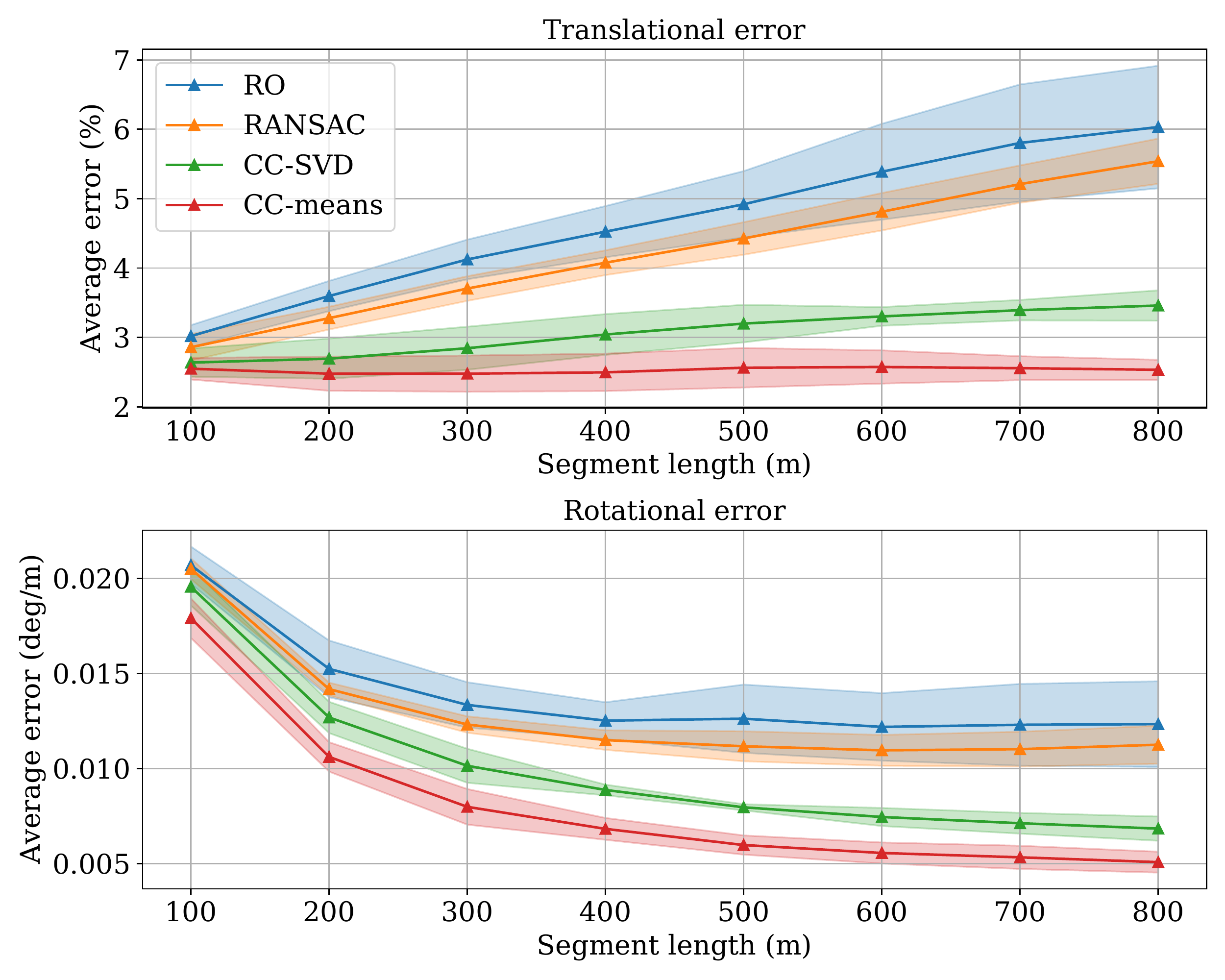}
% \vspace*{-5mm}
\caption{
Mean segment errors with corresponding standard deviation envelopes over the course of the first five sequences of the Oxford \SI{10}{\kilo \metre} route showing translational and rotational error performance for all four methods.}
\label{fig:segment-metrics}
\vspace{-.4cm}
\end{figure}

% \vspace*{.8cm}
\section{Results}
\label{sec:results}
\begin{table}
\centering
\begin{threeparttable}
\caption{Error metrics for ego-motion estimation methods averaged over all tested sequences}
\label{tab:metrics}
\begin{tabular}{@{}lccc@{}}
\toprule
Method           & Tr. [\SI{}{\percent}] & Rot. [\SI{}{\degree}/\SI{}{\metre}] \\ \midrule
RO (full matches)                 & 4.6755             & 0.0139                       \\
RO w/ RANSAC      & 4.2379             & 0.0129               		\\
RO w/ CC*, SVD     & 3.0717             & 0.0101               		\\
RO w/ CC*, means   & \textbf{2.5287}    & \textbf{0.0082}              \\

\bottomrule
\end{tabular}
\begin{tablenotes}[flushleft]
\item []*Constant-curvature method
\end{tablenotes}
\end{threeparttable}
\end{table}

% RO                  & 4.4096             & 0.0137                       \\
% RO with RANSAC      & 4.3426             & 0.0135               		\\
% RO with CC*, SVD     & 2.5517             & 0.0086               		\\
% RO with CC*, means   & \textbf{2.3218}    & \textbf{0.0076}              \\
Over the five \SI{10}{\kilo\meter} test sequences, trajectory metrics are summarised in \Cref{tab:metrics} for each method's performance.
\Cref{tab:fullmetrics} details the performance of the four different systems on each of these five sequences to provide a more in-depth set of results.
% While the full set of metrics over each of the five \SI{10}{\kilo\meter} test sequences we select is detailed in~\Cref{tab:fullmetrics}, average trajectory metrics over segment lengths are summarised in \Cref{tab:metrics} and shown graphically in \Cref{fig:segment-metrics} for each method's average performance.
% across the first five \SI{10}{\kilo\meter} sequences from the dataset.

\fixM{\label{2.2.1}As expected, because of its widespread application to outlier removal}{R2.2.1}, the na\"ive RANSAC method \fixS{only}performs\fixS{marginally} better than when using the full set of matches.
\fixM{\label{2.2.2}However, this improvement is marginal.
Therefore, the implausible associations in~\cref{fig:matches_problem} are evidently not dealt with effectively by standard RANSAC.
% }{R2.2.2}
%Setting the RANSAC threshold parameter is difficult, as if it is too low the estimates are almost identical to the full match set, and when too high we observed poorer odometry performance due to too many valid matches being discarded.
We expect that this is because}{R2.2.2} the initial match selection performed by \acrshort{ro} has already selected matches based on their plausibility from a geometric standpoint, which is similar to what this baseline method is doing. 
However, when leveraging the constant-curvature constraint to select matches, we see a much larger drop in both translational and rotational error.
For a direct comparison between methods, the \acrshort{svd} algorithm is used to compute poses on the refined subset and demonstrates the quantitative benefit of removing spurious matches.

Further to this, estimating pose from a single landmark association allows the constant-curvature motion parameters to be converted into their $\mathbb{SE}(2)$ form for every inlier, where we can find the mean of $\Delta x$, $\Delta y$, and $\Delta \theta$.
This outperforms the \acrshort{svd} approach, as it reduces the impact of relatively poor associations more so than the \acrshort{svd} algorithm, which assigns all members of the selected subset an equal weighting.

Taking a closer look at the estimates, we see that the lateral error that is present when we originally use the full set of matches (i.e. \textit{RO (full matches)} in~\Cref{tab:metrics,tab:fullmetrics}) is erratic while our proposed system is steady in its estimates, particularly in areas when there should be smooth forward motion i.e. no lateral displacement, shown in~\Cref{fig:lateral_motion_problems}.
\fixM{\label{2.3}During sharp turns, there is some underestimation of the lateral motion likely due to the applied quantile limits selecting a narrower set of associations as compared to more common gradual turns or straight trajectories.
% There is some underestimation of the lateral motion.
% However, this only occurs during sharp turns, and is likely due to the applied quantile limits selecting a narrower set of associations as compared to gradual turns or straight trajectories.
While being a limitation of our system, this method still tracks the ground truth signal \textit{far} closer than the baseline system (see~\cref{tab:fullmetrics} and other results in~\cref{sec:results}), which \textit{itself} sometimes over- or underestimates the true motion and presents highly deviant estimates when the vehicle is travelling straight forwards.}{R2.3}
Stated plainly, the ego-motion estimates that our proposed system produces are more plausible given the real constraints that the vehicle is under -- i.e. cars do not suddenly move sideways.

Aggregating over all segment lengths for each sequence, we see from~\cref{tab:fullmetrics} translational error in the range of \SIrange{2.25}{2.84}{\percent} as compared to \SIrange{4.26}{5.55}{\percent} for the baseline RO system, a reduction in error in the range \SIrange{51}{53}{\percent}.
Similar gains are made for rotational errors.

\begin{table*}
\centering
\begin{threeparttable}
\caption{Error metrics for ego-motion estimation methods over all experiences from \texttt{2019-01-10}}
\label{tab:fullmetrics}
\begin{tabular}{@{}lccccccccccccccc@{}}
\centering
% \toprule
Sequence $\rightarrow$           & \multicolumn{2}{c}{\texttt{11-46-21}}           & \multicolumn{2}{c}{\texttt{12-32-52}}           & \multicolumn{2}{c}{\texttt{14-02-34}}           & \multicolumn{2}{c}{\texttt{14-50-05}}           & \multicolumn{2}{c}{\texttt{15-19-41}} \\
\toprule
Method $\downarrow$          & Tr. [\SI{}{\percent}] & Rot. [\SI{}{\degree}/\SI{}{\metre}]         & Tr. [\SI{}{\percent}] & Rot. [\SI{}{\degree}/\SI{}{\metre}]         & Tr. [\SI{}{\percent}] & Rot. [\SI{}{\degree}/\SI{}{\metre}]         & Tr. [\SI{}{\percent}] & Rot. [\SI{}{\degree}/\SI{}{\metre}]         & Tr. [\SI{}{\percent}] & Rot. [\SI{}{\degree}/\SI{}{\metre}] \\ \midrule
RO (full matches)                 & 4.54             & 0.013              & 5.55             & 0.017              & 4.26             & 0.013              & 4.44             & 0.014              & 4.61             & 0.013                       \\
RO w/ RANSAC      & 4.50             & 0.013     & 4.20             & 0.012     & 4.09             & 0.013     & 4.34             & 0.013     & 4.05             & 0.012               		\\
RO w/ CC*, SVD     & 2.96             & 0.010    & 3.15             & 0.010    & 3.32             & 0.011    & 2.92             & 0.010    & 3.32             & 0.011               		\\
RO w/ CC*, means   & \textbf{2.35}    & \textbf{0.008}      & \textbf{2.57}    & \textbf{0.009}      & \textbf{2.63}    & \textbf{0.008}      & \textbf{2.25}    & \textbf{0.007}      & \textbf{2.84}    & \textbf{0.009}              \\

\bottomrule
\end{tabular}
\begin{tablenotes}[flushleft]
\item []*Constant-curvature method
\end{tablenotes}
\end{threeparttable}
\end{table*}

% Further to this, estimating pose from a single landmark association allows the constant-curvature motion parameters to be converted into their $\mathbb{SE}(2)$ form for every inlier, where we can find the mean of $\Delta x$, $\Delta y$, and $\Delta \theta$.
% This outperforms the \acrshort{svd} approach, as it reduces the impact of relatively poor associations more so than the \acrshort{svd} algorithm, which assigns all members of the selected subset an equal weighting.
When looking across all five sequences, our contribution (\textit{CC-means}) has been able to reduce absolute error in translation by approximately \SI{2.15}{\percent}, almost halving the original error in \gls{ro}.
Rotational error was just \SI{0.0082}{\degree\per\metre}, down from \SI{0.0139}{\degree\per\metre} when using all matches.

\begin{figure}
\centering
\includegraphics[height=8cm,keepaspectratio]{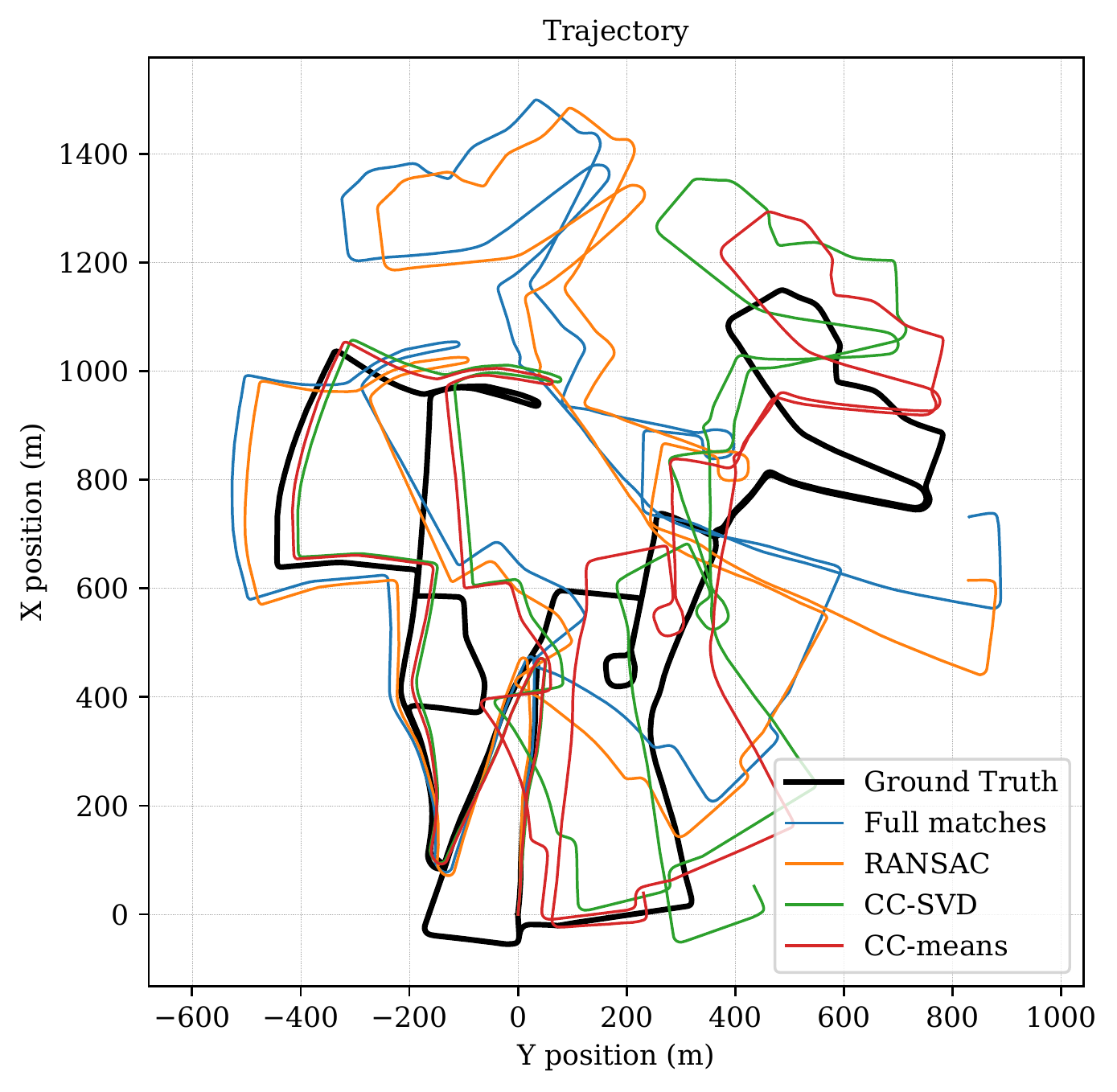}
\caption{
% \gadd{all matplotlib - slightly bigger legend text}
The ground traces for the four methods from the \texttt{2019-01-10-15-19-41} sequence show accumulated odometry performance over time as compared to optimised ground truth. \fix{This figure is best viewed in colour.}}
\label{fig:ground_trace}
\vspace{-.6cm}
\end{figure}

\Cref{fig:ground_trace} \fixM{\label{5.5}(best viewed in colour)}{R5.5} shows the overall trajectories from each method compared with ground truth, which provides more of a qualitative result.
Here, the reduction in drift over kilometres of driving can be exemplified by paying attention to the square loop located at approximate coordinates $(\SI{200}{\metre};~\SI{400}{\metre})$ in the ground truth trace shown in black.
This section of the route occurs near to the end of the \SI{10}{\kilo\meter} sequence, where the majority of any accumulated error is observable.
At this point, our proposed method has drifted approximately \SI{140}{\metre} with a few degrees of yaw error, while the original system is more than \SI{450}{\metre} off-course and has a yaw estimate more or less perpendicular to the ground truth.

Making direct comparisons with other \acrshort{ro} systems is difficult as they publish their results on different sets of either 7 or 8 of the 32 sequences from the \textit{Oxford Radar Robotcar Dataset}~\cite{barnes2020oxford}, as there is no universally recognised ``test set''.
However, the translational error can provide a loose comparison metric, despite being reported on different sequences from the dataset.
In achieving \SI{2.53}{\percent}, our proposed system brings the baseline \gls{ro}~\cite{cen2018} on which it is applied (having been relegated with \SI{4.68}{\percent}) back into competition with more recently developed top-performing \gls{ro} systems, which include: RadarSLAM~\cite{hong2020radarslam} (\SIrange{3.11}{2.21}{\percent}), Barnes' Under The Radar~\cite{barnes2020under} (\SI{2.06}{\percent}), HERO~\cite{burnett2021radar} (\SI{1.99}{\percent}), CFEAR~\cite{adolfsson2021cfear} (\SI{1.76}{\percent}), and Barnes' Masking by Moving~\cite{barnes2020masking} (\SIrange{2.78}{1.16}{\percent}).
This is achieved without any use of learning techniques that introduce additional complications, like those required in Under The Radar~\cite{barnes2020under}, HERO~\cite{burnett2021radar}, and Masking by Moving~\cite{barnes2020masking}.
Furthermore, our new method is applicable to any point-based \gls{ro} pipeline (e.g. CFEAR~\cite{adolfsson2021cfear}), where it should improve match refinement while introducing relatively negligible additional processing overhead.
% As a refinement step in a point-based \gls{ro} pipeline, it should improve other point-based \gls{ro} implementations like most of those mentioned above which are not making use of this information.
% Given the simplicity of this refinement step, it would introduce relatively little additional processing overhead, making it an attractive option for use in future implementations.
%\roberto{In addition to running more Oxford 10k data, possibly extend to evaluate on non-Oxford radar data, like ``MulRan: Multimodal Range Dataset for Urban Place Recognition'' and ``RADIATE: A Radar Dataset for Automotive Perception'' - there's no ground truth for these... Oxford 10k is the odometry benchmark for radar.}

\section{Conclusion}
\label{sec:conclusion}
This work shows that, by considering the physical constraints on how a robot is able to move through its environment, we can drastically reduce both translational and rotational error in \gls{ro}.
Crudely filtering out matches based on the overall consensus of the group is potentially helpful, but ultimately a na\"ive approach to this problem.
Instead, if we consider matched landmarks based on how an appropriately-constrained robot would have had to move in order to generate those associated candidates, the match refinement task becomes far more natural and yields compelling improvements in ego-motion estimation at a low cost.

% stress how light-weight this method is - it's two equations, nothing complicated, and then we sort and take a mid-section. It's very cheap.
Specifically, this approach has the advantage of being extremely light-weight: finding the two parameters that describe ego-motion for each match requires two simple equations, before a sorting and subset-selection step refines the set of matches.
Solving for the final pose estimate then just involves taking a mean for $\Delta x$, $\Delta y$, and $\Delta \theta$ which is also inexpensive.

% we use a strong prior, it's an obvious thing to take advantage of the platform constraints and no one is doing it. No need to try and train something to learn this from hundreds of km's of driving data
Our contribution makes use of a strong prior on how our vehicle is constrained to move.
By folding this information into our system at the match-selection level, we see reduced lateral errors introduced by poor matches that adversely affect odometry estimates.
This is achieved without the need to collect hundreds of kilometres of training data which is usually required to learn how a vehicle is likely to move.

% its adaptable to other point based ROs, and even lidar if you like, as we just work with ranges and bearings
Lastly, our method is portable: any point-based odometry system that uses a range sensor (radar, LiDAR) to measure ranges and bearings to landmarks in the environment, captured from a non-holonomic vehicle, should benefit from our contribution presented here.

%\roberto{It's also super fast and basically for free. Just plug-and-play with any data association system you like.}

\section*{Acknowledgements}

We are grateful to our partners at Navtech Radar, the Assuring Autonomy International Programme -- a partnership between Lloyd’s Register Foundation and the University of York -- and EPSRC Programme Grant ``From Sensing to Collaboration'' (EP/V000748/1).

\bibliographystyle{IEEEtran}
\bibliography{biblio}

\appendix
\label{sec:appendix}
\setcounter{equation}{0}
\renewcommand{\theequation}{A.\arabic{equation}}
As referenced in \Cref{subsec:single-landmark}, given the ranges $r_0$, $r_1$ and bearings $\beta_0$,  $\beta_1$ we can find $\theta$ and $R_{ICR}$.
The diagram has the landmark on the left of both relative poses to avoid overcrowding of symbols on the right-hand side, but this derivation is clearer if we avoid reflex angles.
As such, we define angles $\phi_0 = \beta_0 - 2\pi$ and $\phi_1 = \beta_1 - 2\pi$ before getting started.
Firstly, we express $\gamma$ in terms of $\phi_1$ and $\theta$ as
\begin{equation}
	\gamma = \phi_1 - \theta
\end{equation}
Then find $\alpha$ by summing angles in the constructed triangle $\Delta P G_{t-1} G_t $ and rearranging:
\begin{align*}
	\pi &= \alpha + \gamma + \sfrac{\theta}{2} \\
	\therefore \alpha &= \pi - \phi_1 + \sfrac{\theta}{2} \label{eq:alpha} \numberthis
\end{align*}
Then, by the sine rule, we can express
\begin{align*}
	\frac{r_0}{\sin \alpha} &= \frac{r_1}{\sin \left(\phi_0 + \sfrac{\theta}{2} \right)} \\
	\frac{r_0}{\sin \left(\pi - \left(\phi_1 - \sfrac{\theta}{2}\right)\right)} &= \frac{r_1}{\sin \left(\phi_0 + \sfrac{\theta}{2} \right)} \\
	\frac{r_0}{r_1} &= \frac{\sin \left(\phi_1 - \sfrac{\theta}{2}\right)}{\sin\left(\phi_0 + \sfrac{\theta}{2}\right)}
\end{align*}
Expanding using the trigonometric identity $\sin(A+B)=\sin A \cos B + \cos A \sin B$ gives
\begin{equation*}
	\frac{r_0}{r_1} = \frac{\sin(\phi_1)\cos\left(-\sfrac{\theta}{2}\right)+\cos(\phi_1)\sin\left(-\sfrac{\theta}{2}\right)}{\sin\left(\sfrac{\theta}{2}\right)\cos(\phi_0) + \cos\left(\sfrac{\theta}{2}\right)\sin(\phi_0)}
\end{equation*}
and dividing all terms through by $\cos\left(\sfrac{\theta}{2}\right)$
\begin{equation*}
	\begin{split}
		\sfrac{r_0}{r_1}\left(\tan\left(\sfrac{\theta}{2}\right)\cos(\phi_0)+\sin(\phi_0)\right) \\ = \sin(\phi_1) - \cos(\phi_1)\tan\left(\sfrac{\theta}{2}\right)
	\end{split}
\end{equation*}
then moving $\theta$-terms to the left-hand side
\begin{equation*}
	\begin{split}
		\sfrac{r_0}{r_1}\tan\left(\sfrac{\theta}{2}\right)\cos(\phi_0) + \cos(\phi_1)\tan\left(\sfrac{\theta}{2}\right) \\ = \sin(\phi_1) - \sfrac{r_0}{r_1}\sin(\phi_0)
	\end{split}
\end{equation*}
and finally rearranging to express $\theta$ as
\begin{equation}
	\theta = 2\arctan\left( \frac{\sin\phi_1 - \sfrac{r_0}{r_1}\sin\phi_0}{\sfrac{r_0}{r_1}\cos\phi_0+\cos\phi_1}\right)
\end{equation}

\begin{figure}%[!h]
	\centering
    \includegraphics[height=6.5cm,keepaspectratio,trim={24cm 4.5cm 9cm     3cm},clip]{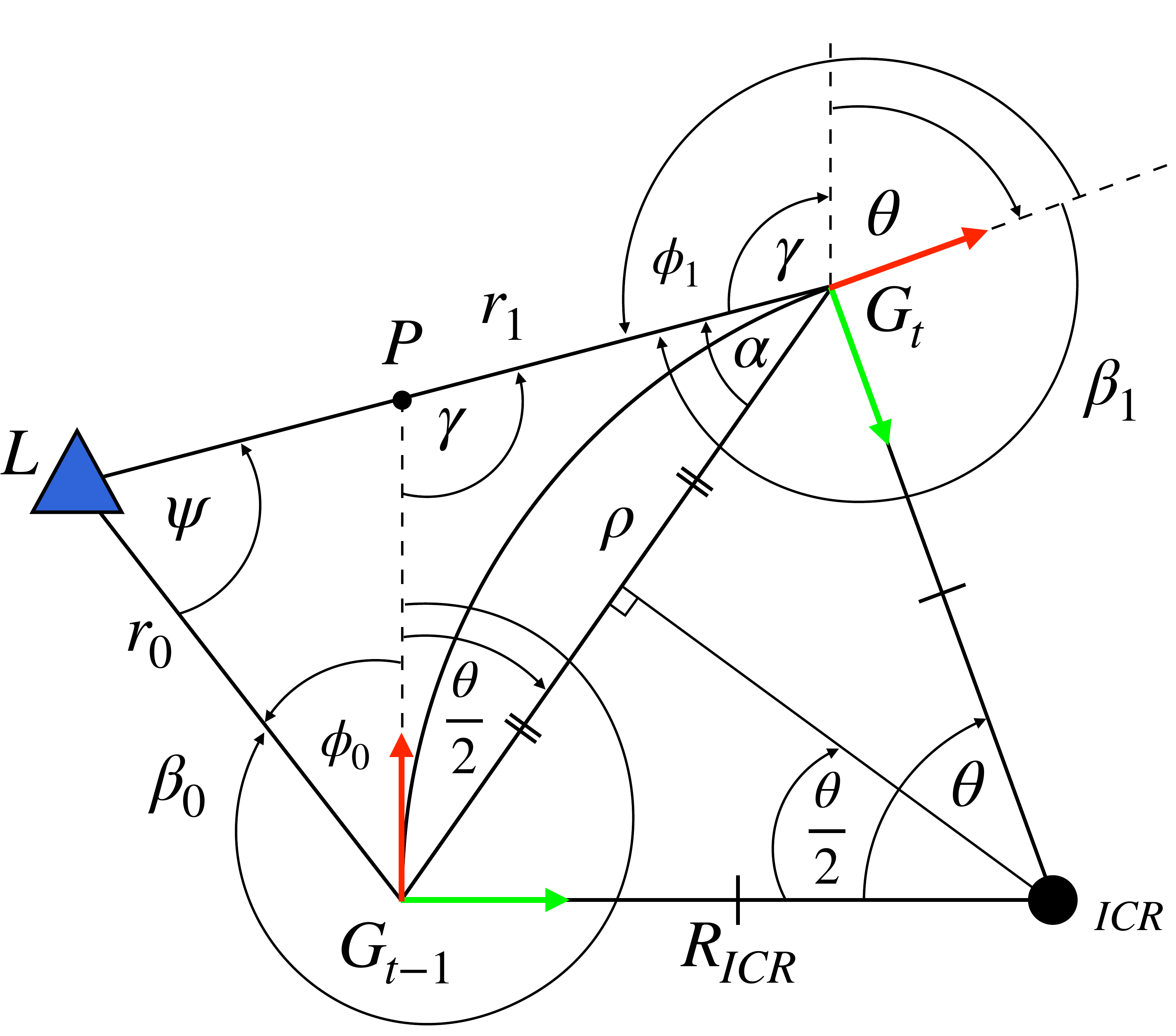}
	\caption{As in \Cref{fig:cm-diagram}, the robot observes a landmark $L$ from two successive poses.
	Included here are additional labels required to step through the derivation of the constant-curvature motion equations.}
	\label{fig:derivation-diagram}
\vspace{-.5cm}
\end{figure}

The diagram in \Cref{fig:derivation-diagram} shows that we can express $\Delta x$ and $\Delta y$ in terms of the trigonometric components of $\rho$, which is computable given ranges $r_0$, $r_1$ and angles $\phi_0 = -\beta_0$ and $\phi_1 = -\beta_1$ as before.
Firstly, we find angle $\psi$ between the start and end pose from the landmark's position at $L$ by summing angles in the triangle $\Delta L G_{t-1} G_{t} $ and using \Cref{eq:alpha} to give
\begin{align*}
	\pi &= \alpha + \psi + \phi_0 + \sfrac{\theta}{2} \\
	\therefore \psi &= \phi_1 - \phi_0 - \theta 	\label{eq:psi} \numberthis
\end{align*}
and relating $\rho$ to $R_{ICR}$ as %TODO - pick better name for R_{ICR}...
\begin{equation}\label{eq:rho-to-R}
	\rho = 2R_{ICR} \sin\left(\sfrac{\theta}{2}\right)
\end{equation}
Then using the sine rule, we have
\begin{align*}
	\frac{\rho}{\sin \psi}&=\frac{r_1}{\sin \left(\phi_0+\sfrac{\theta}{2}\right)} \\
	\rho &= \frac{r_1 \sin \psi}{\sin \left(\phi_0+\sfrac{\theta}{2}\right)}
\end{align*}
and substituting in \Cref{eq:rho-to-R,eq:psi} to solve for $R_{ICR}$
\begin{align*}
	2R_{ICR} \sin\left(\sfrac{\theta}{2}\right) &= \frac{r_1 \sin \psi}{\sin \left(\phi_0+\sfrac{\theta}{2}\right)} \\
	\therefore R_{ICR} &= \frac{r_1\sin(\phi_1 - \phi_0 - \theta)}{2\sin\left(\sfrac{\theta}{2}\right)\sin\left(\phi_0 + \sfrac{\theta}{2}\right)} \numberthis
\end{align*}
Lastly, expressing $\theta$ and $R_{ICR}$ in terms of our bearings $\beta_0$ and $\beta_1$ after the above derivation yields:
\begin{align}
	\theta &= 2\arctan\left(\frac{\sfrac{r_0}{r_1}\sin \beta_0 - \sin \beta_1}{\sfrac{r_0}{r_1}\cos \beta_0 + \cos\beta_1}\right)\\
	R_{ICR} &= \frac{r_1\sin(\beta_0 - \beta_1 - \theta)}{2\sin\left(\sfrac{\theta}{2}\right)\sin\left(\sfrac{\theta}{2} - \beta_0 \right)}
\end{align}

%=================================================================

\ifthenelse{\boolean{review}}{
\clearpage
\newpage
\section*{Author Response to Reviewer 2}

% Contribution of the Paper 
% =========================
% 
% This paper extends a previously published Radar Odometry system which
% is based on associating Radar landmarks in successive scans and finding
% the geometric rigid body transformation in SE2 that best explains the
% geometry of all associations. 
% 
% The contribution of this paper is a new outlier rejection scheme which
% is based on two assumptions: 1) A car-like nonholonomic kinematic
% motion model, from which it follows that the vehicle’s motion can be
% described as driving on a circle with certain curvature, and 2) that
% the curvature between two successive radar scans is constant. 
% 
% The authors show that using these assumptions, one landmark associated
% between two scans is enough to fully describe the vehicle’s motion thus
% constrained by the circle’s radius and angle. The authors sort
% associations by angle, select a certain range of associations around
% the median angle as inliers, and disregard the remainder as outliers. 
% 
% They then go on to show that the Radar Odometry position error is
% almost halved compared to that of the baseline without outlier
% rejection, and another baseline with RANSAC for outlier rejection. 
% 
% 
% 
% Organization and Style 
% ======================
% 
% The paper is well organized and written in an understandable language.
% It is easy to follow without hiding technical details. I feel confident
% I could reproduce the proposed extension, given the paper alone. 
% 
% 
% 
% Technical Accuracy 
% ==================

The authors would like to thank the reviewer for their thorough comments and useful observations.
The authors have addressed all of these concerns in the resubmitted version of the paper and summarised them here.
Alongside the resubmitted manuscript, the authors have provided an annotated version in which all changes have been highlighted and marked with a unique ID to help the reviewers locate the portion of the work that addresses their concerns.
\\
\\
%-----------------------------------------------------
%\checkmark
\OB{2.1}~\textit{
% The paper does an excellent job of describing the proposed approach and proving its advantage. 
I would still make the two core assumptions mentioned above more explicit and briefly mention when they are valid and when they reach their limit. (By the way: How do they hold up for the dataset used?).
Also, the “Constant-curvature Motion Constraint” should be more clearly defined \ldots I first assumed it meant there is a constant curvature constraint between two consecutive scan-matching results.
}\\

Please see \REP{2.1.1} on \cpageref{2.1.1} where we have added text to make the assumption more explicit.
Please see \REP{2.1.2} and \REP{2.1.3} on \cpageref{2.1.2} where we have amended the existing text to clarify why our modelling assumptions are appropriate for the motion of the vehicle in the dataset that we use for our experiments.
Please also see \REP{2.1.4} on~\cpageref{2.1.4} where we make it clear that a window of scan matching results is not used in enforcing our constraint.
\\
\\
% The assumptions referred to relate to the nonholonomic motion model used, and the assumption that the curvature of a path traversed between two poses is constant.
% In (cite sentence here) it is stated that the selected model ``serves as a suitable approximation for the motion of our vehicle in an urban environment where wheel slippage is minimal'', which outlines where this assumption holds, while also mentioning its limitation when used in settings where there may be significant wheel slippage.
% Given that the dataset used is in an urban setting, this also answers the appropriateness of this model in this context.
% The assumption that curvature is constant between poses where radar scans were captured is made based on the reasonably short duration (0.25 s) between scans.
% This has been added into the text to make this assumption clear, please see page~\pageref{fix:short-duration}.
% \\
% \\
% %-----------------------------------------------------
% %\checkmark
% \OB{2.2}~\textit{Also, the “Constant-curvature Motion Constraint” should be more clearly defined.
% From the paper title alone, I first assumed it meant there is a constant curvature constraint between two consecutive scan-matching results.
% But in fact, this is only true between two consecutive scans, i.e., for a single scan-matching result. 
% }\\

% This constraint is applied between two poses, and is already described in detail in the relevant section~\cref{subsec:motion-model}.
%-----------------------------------------------------
%\checkmark
\OB{2.2}~\textit{Figure 3 contains some obvious outlier associations that would easily be removed by a RANSAC scheme.
But then at the beginning of Section VI, the paper says: “This is unsurprising, as the initial match selection performed by RO has already selected matches based on their plausibility from a geometric standpoint, which is similar to what this baseline method is doing.” Why do these implausible associations then still show up in Figure 3?
}\\

Please see \REP{2.2.1} and \REP{2.2.2} on~\cpageref{2.2.2} where we have amended the text.
Please see \REP{2.2.3} on~\cpageref{2.2.3} where we have explained why false matches propagate through our baseline system and stated that while you might expect the false matches to be dealt with by RANSAC, they are evidently not (by our results), as RANSAC also does not exploit domain knowledge of how the vehicle motion is constrained.
\\
\\
%-----------------------------------------------------
%\checkmark
\OB{2.3}~\textit{In Figure 2 it seems like lateral motion is systematically underestimated by a factor of 2, which does not look good at all.
Do you have an explanation for that?
}\\

While the main focus of our work is on the highly deviant estimates during forward motion, and the overall performance boost is nevertheless significant, this is a good observation and worth stating in the paper in reference to the limitations of our system.
Please see \REP{2.3} on~\cpageref{2.3} for the amended text, where we explain the likely reason.
% \Cref{fig:lateral_motion_problems} shows the lateral motion for our baseline RO system, our contribution in this paper, and a ground truth signal.
% The underestimation referred to in the review comment only occurs during sharp turns, and is likely due to the selected quantile limits selecting a narrower set of associations as compared to gradual turns or straight trajectories.
% While our system is not perfect, it still clearly tracks the ground truth signal more closely than the baseline system, which is made clear in \Cref{fig:lateral_motion_problems}.

% Sharp turns seem to be problematic, the lateral estimates are nearer to GT when turning is less erratic (around index = 650ish). Possibly due to the quantile selection being too strict, where a very narrow band is used during these manoeuvres. It’s still far better than RO, but it’s not a perfect system. (would be good to add more here to explain this result) 

% Summary 
% =======
% 
% This is a well-written paper about an incremental improvement to an
% existing Radar Odometry pipeline. The Constant-Curvature Constraint as
% proposed here seems generic enough for it to be applied to other
% systems based on associating landmarks between consecutive scans.
% 
% One could argue that the proposed approach is simple and question the
% innovativeness of the paper. To me, however, its simplicity is what
% makes it so attractive: As far as I understood the only parameter is
% the quantile range. 

\clearpage
\newpage
\section*{Author Response to Reviewer 5}

% Overview:
% This paper presents a pipeline that leverages robot dynamics to
% geometrically constrain 1-point landmark associations in radar-based
% odometry. The dynamics in this paper assume a constant-curvature model
% in a non-holonomic robot, and given landmark range and bearing at two
% different time instances, in addition to the radius of curvature of the
% robot, a relative pose estimate is estimated. Contributions of this
% work include developing a method for the 1-point RANSAC model by
% Scaramuzza et. al in radar (range-based) odometry, and a method for
% outlier rejection through a conforming subset of landmark matches. The
% paper well-written, overall presentation is good, and experiments show
% that the proposed method works well.

The authors would like to thank the reviewer for their thorough comments and useful observations.
The authors have addressed all of these concerns in the resubmitted version of the paper and summarised them here.
Alongside the resubmitted manuscript, the authors have provided an annotated version in which all changes have been highlighted and marked with a unique ID to help the reviewers locate the portion of the work that addresses their concerns.
\\
\\
%-----------------------------------------------------
%\checkmark
\OB{5.1}~\textit{
My primary concern is in regards to the framing around the novelty of the contributions. Using vehicle dynamics for single-landmark odometry is a well-known result in the vision community which was first presented by Scaramuzza et. al [1], and the proposed paper seems to be just a translation of that seminal work to radar modality. While this translation in itself is novel -- as the method has not been employed to radar (range) modalities -- and the corresponding theoretical development using range and bearing measurements is new, the paper reads a little similarly to Scaramuzza et. al's work. The emphasis on contributions may also be slightly misleading for those who are not familiar with the previous result from the vision community. I would suggest further differentiating between the two works (i.e., not just stating that it's different sensing modalities, and more emphasis on the conforming subset method) to strengthen the contributions of the work.
To further expand on my comment above (and to perhaps provide a few suggestions), I think it's worth considering a comparison between the two methods in the Experiments section. In other words, how do the resulting trajectories compare between using Scaramuzza et. al's vision-based method and the proposed radar-based method? This would provide the community valuable insight in how radar stacks up against vision. I would also like to see how, quantitatively, the proposed "comforming-subset" outlier rejection method compares against their histogram-based voting method (i.e., Fig. 8). 
}\\

% Did we only say it was different sensor modalities?
In \cref{sec:related}, we emphasise that the work of Scaramuzza et al.~\cite{scaramuzza2009real} is perhaps the closest related work in this area, in that it too uses knowledge of the underlying vehicle dynamics to discard outlier associations.
We also state that although the two works take a similar approach to solving such a problem, the details of the theory ``differ significantly due to the fundamental differences between sensor modalities'', which already strongly states what this comment is requesting.
Furthermore, in \cref{subsec:motion-model} we expand on why we cannot directly apply the earlier work of Scaramuzza et al.~\cite{scaramuzza2009real} to the context of a range-bearing sensor like radar, and detail the differences between that method and ours which is presented here.
Finally, we also discuss the conforming subset method and how it was chosen after the final pose estimation method from Scaramuzza et al.~\cite{scaramuzza2009real} yielded poor results in this context in \cref{subsec:final-pose-estimation}.
\\
\\
However, in light of this comment, we amend the text in~\REP{5.1.1} on~\cpageref{5.1.1} to further discuss the differences to~\cite{scaramuzza2009real}.
We point out that our derivation results in a different set of equations, specific to the geometry of this class of sensor, in~\REP{5.1.2} on~\cpageref{5.1.2}.
\\
\\
We consider the comparison between vision and sensor modalities beyond the scope of this work, but refer interested readers to existing work in that area -- see \REP{5.1.3} on~\cpageref{5.1.3}.
\\
\\
We comment further on why we avoided the histogram method of [17] and proposed a conforming subset, see \REP{5.1.4} on~\cpageref{5.1.4}.
\\
\\
% \\
% \\
% %-----------------------------------------------------
% %\checkmark
% \OB{5.2}~\textit{
% }\\

% This request would be a significant deviation from the core contribution in this paper.
% There are existing works that go into detail on radar vs vision in the ego-motion estimation task (see \cite{cen2018}, \cite{hong2020radarslam} already cited in this paper) and running a larger experiment like this arguably obscures the contribution of our new sub-component.
% It would be difficult to assess the performance of the new system, as results would conflate any real differences with those that are attributable to the sensor modality in use.
% % Would such a paper be about a new method, or is it a vision vs radar experiment? It seems like it would become two papers, and I think it’s safe to say it is out of scope.
% The comment on using the conforming subset vs histogram voting was something we assessed in the preliminary experiments during development, which determined the conforming subset method was clearly superior, as mentioned in \cref{subsec:final-pose-estimation}.
% % Space constraints (and not going into unnecessarily fine detail) mean we can’t cover all aspects of the entire development of the system.
%-----------------------------------------------------
%\checkmark
\OB{5.2}~\textit{
Fig. 1 is hard to read; consider increasing font size and zooming into a subset for readability.
% Also consider a different "front-page" image.
}\\

\Cref{fig:sysoverview} has been amended for text size to make it easier to read.
Please see \REP{5.2.1} on~\cpageref{5.2.1} where we direct the reader to a zoomed-in view in a later figure.
% It is unclear what the reviewer meant by using a different ``front-page'' image.
% This figure was chosen to represent the essence of the contribution which it encapsulates in a simple form.
\\
\\
%-----------------------------------------------------
%\checkmark
\OB{5.3}~\textit{
Fig. 2 seems like it's out of place and belongs more in the Methods section (rather than right above Related Works).
}\\

This has been adjusted so that \cref{fig:lateral_motion_problems} (previously Fig. 2) now appears in a more logical place in the paper.
\\
\\
%-----------------------------------------------------
%\checkmark
\OB{5.4}~\textit{
A figure which provides the reader with more insight into the overall archtecture would help tremendously.
}\\

% The preliminaries are discussed in detail in \cref{sec:preliminaries} where each component of the baseline \gls{ro} pipeline is presented.
% More detail on the system architecture is contained in the original work which has been cited several times~\cite{cen2018}.
The pipeline described in~\cref{sec:preliminaries}, our baseline system, is from~\cite{cen2018}.
We have amended the text to refer the interested reader to the most helpful figure in that publication, please see~\REP{5.4} on~\cpageref{5.4}.
% We are not presenting a new architecture, so it does not seem relevant to include this figure in addition to the pipeline that is already discussed in depth.
\\
\\
%-----------------------------------------------------
%\checkmark
\OB{5.5}~\textit{
Fig. 8 is a little cluttered; consider using different line styles (e.g., dashed lines, dotted lines, etc.) for different methods in addition to the colors to help with readability.
}\\

We considered this, but the figure would be more visually confusing.
We have, however, amended the text to advise the reader to view/print in colour for best readability -- see Fig. 8 caption and \REP{5.5} on~\cpageref{5.5}.
% Showing a qualitative result like that in \cref{fig:ground_trace} for four systems alongside a ground truth is tricky.
% Adding different line styles in addition to the already present colour differences is arguably going to make this figure appear busier than it needs to be.

}{}
	
\end{document}